\pdfoutput=1

\documentclass[11pt]{article}

\usepackage[]{acl}

\usepackage{natbib}
\usepackage{graphicx}
\usepackage{tabularx}
\usepackage{soul}
\usepackage{colortbl}
\usepackage{latexsym}
\usepackage{array}
\usepackage{booktabs}
\usepackage{multirow}
\usepackage{amsmath}
\usepackage{amssymb}
\usepackage{graphicx}
\usepackage{enumerate}
\usepackage{diagbox}
\usepackage{enumitem}

\usepackage{color}

\newcommand{\ckbpold}{{{CKBP v1}}}
\newcommand{\ckbpnew}{{{CKBP v2}}}

\newcommand{\tabincell}[2]{\begin{tabular}{@{}#1@{}}#2\end{tabular}} 

\usepackage{microtype}

\usepackage{times}
\usepackage{latexsym}
\usepackage{xcolor}
\usepackage{hyperref}
 \definecolor{darkblue}{rgb}{0, 0, 0.5}
  \hypersetup{colorlinks=true, citecolor=darkblue, linkcolor=darkblue, urlcolor=darkblue}

\usepackage{xstring}

\usepackage{color}

\usepackage[T1]{fontenc}

\usepackage[utf8]{inputenc}

\usepackage{microtype}

\usepackage{inconsolata}

\usepackage{graphicx}

\title{\ckbpnew{}: Better Annotation and Reasoning for \\ Commonsense Knowledge Base Population}

\author{Tianqing Fang$^{1}$\thanks{{} {} Equal Contribution}, Quyet V. Do$^{1}$\footnotemark[1], Zihao Zheng$^{2}$, \\
\textbf{Weiqi Wang$^{1}$, Sehyun Choi$^{1}$, Zhaowei Wang$^{1}$,  Yangqiu Song$^{1}$} \\
$^{1}$Hong Kong University of Science and Technology, Hong Kong SAR, China \\
$^{2}$Harbin Institute of Technology (Shenzhen), China\\
\texttt{\{tfangaa, vqdo, yqsong\}@cse.ust.hk}
}

\begin{document}
\maketitle
\begin{abstract}
Commonsense Knowledge Bases (CSKB) Population, which aims at automatically expanding knowledge in CSKBs with external resources, is an important yet hard task in NLP.
\citet{fang-etal-2021-benchmarking} proposed a CSKB Population (CKBP) framework with an evaluation set \ckbpold{}.
However, \ckbpold{} 
relies on crowdsourced annotations that suffer from a considerable number of mislabeled answers, and the evaluation set lacks alignment with the external knowledge source due to random sampling.
In this paper, we introduce \ckbpnew{}, a new high-quality CSKB Population evaluation set that addresses the two aforementioned issues by employing domain experts as annotators and incorporating diversified adversarial samples to make the evaluation data more representative.
We show that \ckbpnew{} serves as a challenging and representative evaluation dataset for the CSKB Population task, while its development set aids in selecting a population model that leads to improved knowledge acquisition for downstream commonsense reasoning. 
A better population model can also help acquire more informative commonsense knowledge as additional supervision signals for both generative commonsense inference and zero-shot commonsense question answering.
Specifically, the question-answering model based on DeBERTa-v3-large~\cite{he2023debertav} even outperforms powerful large language models in a zero-shot setting, including ChatGPT and GPT-3.5.
\end{abstract}

\section{Introduction}


\begin{figure}[t]
    \centering
    \includegraphics[width=0.9\linewidth]{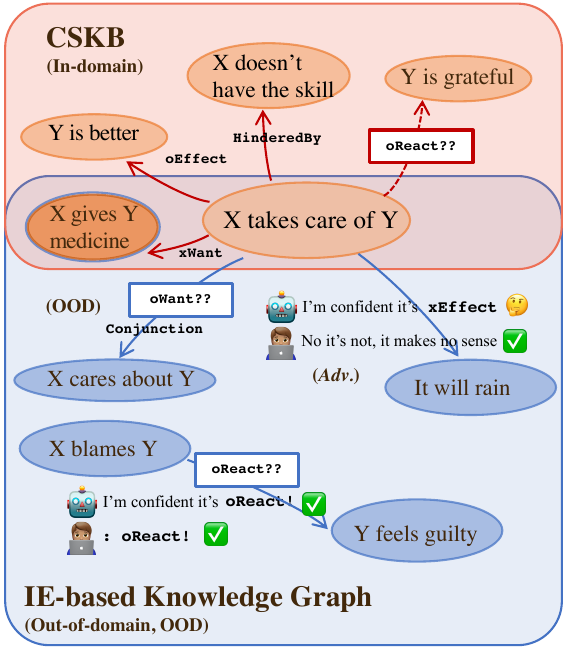}
    \caption{An example of CSKB Population. The coral part indicates the conventional case of CSKB Completion, and the blue part is the population on external knowledge graphs. We include an adversarially constructed sample set in our \ckbpnew{} by re-annotating the confident predictions by language models.}
    \label{fig:intro}
    \vspace{-1em}
\end{figure}

Recently introduced LLMs have shown a remarkable performance on many reasoning benchmarks~\cite{hoffmann2022chinchilla, chowdhery2022palm, bang2023chatgpteval, DBLP:journals/corr/abs-2304-14827}, yet there still exists a need to ensure the alignment between the generation of LLMs with external knowledge at the inference time to avoid hallucination and for safer use~\cite{kim2022soda, he2022rethinking, peng2023check}.
The source of external knowledge, which can be commonsense, factual, or domain knowledge, should be selected and processed carefully depending on the purpose of generation. 
However, existing (high-quality) human-annotated knowledge bases are usually far from complete to serve as the source of external knowledge for LLMs.


Regarding commonsense knowledge bases, to extend limited human annotations, CSKB Population~\cite{fang-etal-2021-benchmarking} stands as a way to acquire missing knowledge, thereby enriching and expanding the existing CSKBs. Unlike CSKB Completion~\cite{li-etal-2016-commonsense, saito-etal-2018-commonsense, Malaviya2019ExploitingSA}, which adopts a close-world assumption and only deals with entities and events within CSKBs, the Population task deals with both existing and unseen entities and events, thus requiring a more generalized reasoning ability.

Several works have been conducted on CSKB Population.
\citet{fang-etal-2021-benchmarking} studied a framework that links four CSKBs, ConceptNet~\cite{DBLP:conf/aaai/SpeerCH17}, ATOMIC~\cite{sap2019atomic}, ATOMIC$_{20}^{20}$~\cite{hwang2021comet}, and GLUCOSE~\cite{mostafazadeh2020glucose}, to a large-scale discourse knowledge base, ASER~\cite{DBLP:conf/www/ZhangLPSL20,zhang2022aser}. 
The resulting knowledge base not only served as the unified source of commonsense knowledge but also was used as the training set to train population models in order to identify missing commonsense knowledge. 
To evaluate models, the authors created an evaluation set (denoted as \ckbpold{}), in which they applied fine-grained rules to select candidate commonsense knowledge from ASER and enlisted human annotators to manually annotate these candidates.

However, there are two major limitations in \ckbpold{}.
First, the quality of \ckbpold{} is limited.
\ckbpold{} instances are randomly sampled from the whole population space, resulting in a low recall of plausible commonsense knowledge due to the noise in candidate discourse knowledge. 
Moreover, as pointed out by \citet{davis2023benchmarks}, current crowdsourced commonsense benchmarks often contain a substantial fraction of incorrect answers, we also find it true for \ckbpold{} after manual inspection.
For example, annotators frequently make mistakes on some subtle relations such as \texttt{xIntent}, which should describe an \textit{intention} instead of a \textit{consequence}.
Second, it's unclear how to leverage populated or expanded commonsense knowledge in CKBP to further improve downstream commonsense reasoning. 
All previous investigations into CKBP stay within the population task itself without generalizing to actual downstream applications.


Therefore, to address the two limitations, this work presents a more high-quality and adversarially constructed evaluation set by expert annotation, 
and a comprehensive pipeline for conducting a series of downstream experiments. The aim is to leverage the new CKBP benchmark effectively and facilitate improved utilization for downstream commonsense reasoning tasks.

Leveraging the existing framework, we build \ckbpnew{} by randomly sampling 2.5k instances from \ckbpold{} and adding 2.5k adversarial instances, leading to a total of 5k instances as an evaluation set. 
These instances are then annotated by experts with substantial expertise in machine commonsense.
Then, we present both intrinsic and extrinsic experiments based on \ckbpnew{}. 
We study the performance of both supervised and semi-supervised task-specific models, together with powerful off-the-shelf language models, such as ChatGPT~\cite{openai2022chatgpt} and Vera~\cite{liu2023vera}, and show that the \ckbpnew{} evaluation set is still challenging even for advanced language models.
Moreover, 
by employing a CSKB Population model that demonstrates satisfactory performance on \ckbpnew{},
we can enrich existing CSKBs with diverse and novel knowledge that significantly benefits downstream reasoning.
We present methodologies and experiments on generative commonsense inference~\cite{bosselut2019comet} and zero-shot commonsense question answering~\cite{DBLP:conf/aaai/MaIFBNO21}, and show that the acquired commonsense knowledge can be valuable augmented data on the original CSKB and lead to improved downstream performance. 
In particular, \ckbpnew{}-preferred population model exhibits better alignment than \ckbpold{} with advancements in generative commonsense inference.

In summary, our contributions are three-fold: First, We introduce a new evaluation benchmark \ckbpnew{} for the CSKB Population task, which addresses the quality issues of its predecessor \ckbpold{}. Second, We launch a pioneer study to use populated commonsense knowledge as additional supervision signals to help downstream commonsense reasoning. Third, We conduct extensive experiments and evaluations with different models on both \ckbpnew{} itself as well as downstream generative commonsense inference and zero-shot question answering. The results show that \ckbpnew{} is still a hard task for language models, and the acquired populated knowledge can improve language models' (zero-shot) commonsense reasoning ability on two downstream tasks across six datasets.

\section{Related Work}\label{sec:appendix_related_works}

In this section, we discuss 1) CSKBs and their role in the era of LLMs and 2) methods and benchmarks for completing and populating knowledge bases in general.

\paragraph{Commonsense Knowledge Bases.}
There are many commonsense knowledge bases\footnote{Here, despite the subtle differences between datasets and knowledge bases, we refer to both as knowledge bases} introduced in the past few years, such as ATOMIC2020~\cite{hwang2021comet}, ComFact~\cite{gao-etal-2022-comfact}, CICERO~\cite{ghosal-etal-2022-cicero}, PIQA~\cite{bisk2020piqa}, Numersense~\cite{lin-etal-2020-birds}. 
Unlike the decades-old knowledge base ConceptNet~\cite{liu2004conceptnet} that only focuses on taxonomic commonsense, these knowledge bases study a broad range of commonsense, including human-event-centric, contextualized, physical, numerical commonsense.

Along with pure-symbolic CSKBs whose knowledge is obtained from corpora and stored in textual format, there is a stream of research that works on developing neural(-symbolic) CSKBs, which are either knowledge models such as COMET~\cite{bosselut2019comet} or symbolic CSKBs built by prompting knowledge from language models, such as ATOMIC$^{10\text{X}}$ ~\cite{WestBHHJBLWC22}, SODA~\cite{kim2022soda}. Although the approach seems highly scalable and seems promising to build more and larger CSKBs, knowledge from neural(-symbolic) CSKBs remains unreliable~\cite{kim2022soda, he2022rethinking, peng2023check} thus often needs to have a robust critic model to filter for good/correct knowledge. 

\paragraph{Reasoning over (Commonsense) Knowledge Bases.}

Regarding conventional knowledge bases like Wordnet~\cite{wordnet} and Freebases~\cite{freebase}, tasks involving completion and population have been well-studied as transductive and inductive link prediction problems in the field of graph neural network~\cite{nips2013transe, Yang2014EmbeddingEA, sun2018rotate, shang2019end, DBLP:conf/www/FangZWSH21}. 
Methods powered by pre-trained language models have also been studied in these tasks thanks to the models' representation power~\cite{yao2019kgbert}. 
In that setting, knowledge instances of the knowledge bases are serialized to a text sequence, which serves as input to LMs such as BERT or RoBERTa.

Specific to CSKB Population task on \ckbpold{}, \citet{fang-etal-2021-benchmarking} proposed KGBertSAGE, a combination of KG-BERT~\cite{yao2019kgbert} and GraphSAGE~\cite{hamilton2017inductive}. The model showed higher performance over baselines yet still suffered from the out-of-domain problem. 
The follow-up works use pseudo-labeling~\cite{fang-etal-2022-pseudoreasoner} and constrained prompting~\cite{DBLP:conf/eacl/DoFDWS24} for solving the problem.

There are also works focusing on using the knowledge from commonsense knowledge bases to help downstream commonsense reasoning such as QA~\cite{DBLP:conf/aaai/BianH0021, DBLP:conf/coling/WangCL00JXLZ24}, including sampling logical queries over CSKBs for better commonsense reasoning~\cite{DBLP:conf/acl/FangCSB24}, reasoning over entailment or abstraction relations~\cite{DBLP:conf/naacl/WangSWFZCLS24, DBLP:journals/corr/abs-2305-14869}, and using knowledge-constrained decoding to guide the generation of LLMs~\cite{DBLP:conf/emnlp/ChoiF0S23}, etc.

\section{Dataset Construction}
\label{section:dataset_construction}

In this section, we introduce the task definition, the preparation of the candidate evaluation set, annotation guidelines, and data analysis. 

\subsection{Task Definition}

The task of CKBP~\cite{fang-etal-2021-benchmarking} is defined as follows. 
Given $G^{C} = \{(h, r, t)|h \in H, r \in R, t \in T\}$ (where $H, R, T$ is the set of head events, relations, and tail events), the graph-like knowledge base formed by aligning a union of commonsense knowledge bases $C$ and a much larger discourse knowledge graph $G$ into the same format; 
the goal of CSKB population task is to learn a scoring function that gives a candidate knowledge triple $(h, r, t)$ higher score if the triple is plausible commonsense.
The training process is formulated as triple classification, with ground-truth positive triples from the CSKB $C$ and negative triples randomly sampled from $G^{C} - C$ with an equal amount. 
The model is then evaluated on a human-annotated evaluation set $E$. 
Here, \ckbpnew{} serves as the evaluation set.

\subsection{Dataset Preparation}
We randomly sampled 2.5k instances from \ckbpold{} and 2.5k adversarial instances to form \ckbpnew{}. 
Instances from \ckbpold{} are sampled so that the ratio of the number of triples between relations remains unchanged.
Meanwhile, the adversarial instances are ones from the candidate knowledge base ASER that the finetuned baseline KG-BERT~\cite{yao2019kgbert} model confidently believes they are plausible, i.e., receives plausibility score $\geq 0.9$.
To ensure the diversity of adversarial instances and hence the evaluation set, we adopt an additional diversity filter using self-BLEU following \citet{WestBHHJBLWC22}. 
The triples annotated as negative are considered \textit{hard negatives} as they are what a standard CSKB Population model would favor.
Note that we only consider instances of 15 relations other than \texttt{general Want/React/Effect}, because most of the triples on the three relations are broken sentences in \ckbpold{}. We also remove samples of these relations in the training set.

\begin{table}[t]
\small
\centering
\renewcommand\arraystretch{1.1}

\begin{tabular}{l|ccc}
\toprule
    & \# Triples & \% Plau. & \% Unseen \\

\midrule
\textbf{split} \\
Dev & 958 & 20.46 & 56.79 \\
Test & 4,048 & 22.06 & 60.43 \\

\midrule
\textbf{instance type} \\
In-Domain & 845 & 34.56 & 43.79\\
Out-of-Domain & 1,653 & 11.92 & 63.37 \\
\textit{Adv.} & 2,508 & 23.92 & 61.12 \\

\midrule
\textbf{relation} \\
xWant & 611 & 22.75 & 54.01 \\
oWant & 239 & 25.94 & 58.18 \\
xEffect & 603 & 29.68 & 55.23 \\
oEffect & 172 & 21.51 & 58.91 \\
xReact & 533 & 20.64 & 51.18 \\
oReact & 183 & 13.66 & 50.70 \\
xAttr & 605 & 23.47 & 52.91 \\
xIntent & 239 & 16.32 & 58.40 \\
xNeed & 378 & 25.66 & 55.37 \\
Causes & 236 & 21.61 & 55.41 \\
xReason & 5 & 40.0 & 30.0 \\
isBefore & 157 & 28.03 & 54.80 \\
isAfter & 182 & 24.73 & 55.40 \\
HinderedBy & 777 & 12.1 & 63.17 \\
HasSubEvent & 86 & 26.74 & 61.04 \\

\bottomrule
\end{tabular}
\vspace{-0.05in}
\caption{Statistics of \ckbpnew{}. \# Triples, \% Plausible, and \% Unseen, respectively, indicate the number of triples in the subset, the proportion of plausible triples after label finalization, and the proportion of nodes that do not appear in the training set.}
\label{table:statistics}
\vspace{-0.1in}
\end{table}

\subsection{Annotation Process}

\paragraph{Setup}
We recruited four human experts for the annotation work. 
The experts are graduate NLP researchers with at least one year of experience working on CSKBs. We randomly divide 5k samples into 4 parts, then for $i$ from 0 to 3, assign the $i^{th}$ and $(i+1 \ \text{mod}\ 4)^{th}$ parts to the $i^{th}$ expert.
In this way, two different annotators annotate each triple, and we can fully compare the pairwise agreement between all four annotators.
Experts are provided with knowledge triples in the format of $(h,r,t)$, referencing the definition and examples of all relations in~\citet{hwang2021comet}. 
We ask annotators to judge the plausibility of triples in a three-point Likert scale with corresponding scores: Always/Often (1), Sometimes (0.5), Rarely/Never/Ambiguous/Invalid (0). 
The final label of an instance is determined as \textit{plausible} if and only if it receives at least one score of 1 and the other score is at least 0.5. 
For remaining cases, the final label is \textit{implausible}. 
After finalizing the annotation, we split the evaluation set into development and test sets with a ratio of 1:4 with the preservation of distribution w.r.t labels, relations, and instance types.
To estimate human performance, we treat expert annotations as two sets of predictions and compare them to the final labels.

Similar to \ckbpold{}, we categorize the evaluation set into three groups based on their origin, which are
1) ID: in-domain, whose head and tail events are all from CSKBs,
2) OOD: out-of-domain, which has at least one event outside of CSKBs (equivalent to ``CSKB head + ASER tail'' and ``ASER Edges'' in \ckbpold{}), and 
3) \textit{Adv.}: adversarial examples newly introduced in \ckbpnew{}.

\paragraph{Quality Control}
Although annotators are experts with a clear understanding of the CSKB Population, we acknowledge the ambiguity of CSKB relations and the difficulty in discriminating between them. To control the quality, 
we provide guidance as a list of scoring criteria. 
We also carried out a dry run, which asked them to annotate 60 instances covering all relations in order to establish a unified understanding of the problem among participants.

After that, we carry out the main round, where the annotators perform their jobs individually and independently. 
Throughout the process, we regularly conduct random checks on the samples and engage in discussions with annotators to address any disagreements. 
We then use the insights gained from these discussions to update and refine our guidance iteratively.
After the individual annotation, we facilitated a conflict resolution session to address instances with contrasting scores of 1 and 0. 
After resolving conflicts, we have the average inter-annotator agreement score IAA as 90.55\%.

\subsection{Data Analysis}

The overall statistics of \ckbpnew{} are shown in Table \ref{table:statistics}. It can be easily observed that the new evaluation set has data imbalance issues. 
However, we do not down-sample the evaluation set to achieve the data balance since the imbalance better reflects the true distribution of plausible and implausible commonsense knowledge in ASER. 
Given this imbalance, we notice that the AUC scores of examined population models will naturally be high. 
Also, in the real application of population models, we focus on the precision and recall of the detection for plausible commonsense instances. 
Thus, in Section \ref{section:experiments}, along with AUC, we also report the binary F1 scores for each experimented model.


\section{Intrinsic Evaluation}
\label{section:experiments}

\begin{table*}[t]
\small
\centering
\renewcommand\arraystretch{1.2}
\begin{tabular}{l|l|cccc|cccc}
\toprule
\multirow{2}{*}{Category} & \multirow{2}{*}{Model} & \multicolumn{4}{c|}{AUC}   & \multicolumn{4}{c}{F1} \\
\cline{3-10} 
& & all & ID & OOD & \textit{Adv.} &  all & ID & OOD & \textit{Adv.} \\
\midrule
\multirow{4}{*}{\tabincell{l}{Zero-shot}}
& GPT2-large & 56.47 & 56.60 & 58.31 & 54.22 &  35.37 & 47.40 & 24.06 & 36.84  \\
& GPT2-XL & 56.79 & 54.47 & 56.70 & 54.63 &  35.22 & 47.62 & 23.49 & 36.65 \\
& GPT3 \scriptsize{\texttt{text-davinci-003}} & 61.63 & 65.93 & 59.17 & 59.98 &  39.44 & 51.09 & 28.57 & 38.20 \\
& ChatGPT \scriptsize{\texttt{gpt-3.5-turbo}} & 65.77 & 70.37 & 62.56 & 62.27 &  45.93 & 62.59 & 44.79 & 26.86 \\
\midrule
\multirow{5}{*}{\tabincell{l}{Supervised\\Learning}}
& KG-BERT (BERT-base) & 71.33 & 84.60 & 64.47 & 62.9 &  45.03 & 69.27 & 26.53 & 41.97 \\
& KG-BERT (RoBERTa-L) & \underline{73.70} & \underline{85.53} & {67.70} & {65.60} &  \underline{46.70} & \underline{69.73} & 30.73 & \underline{43.27} \\ 
& COMET (GPT2-L) & 70.00 & 79.02 & 66.43 & 62.62 &  45.55 & 61.90 & \underline{32.14} & 42.15  \\
& COMET (GPT2-XL) & 70.32 & 79.66 & 66.53 & 63.22 &  45.32 & 63.34 & {31.18} & 40.83 \\
& Vera\* (T5-xxlarge) & 72.45 & 78.84 & \underline{68.40} & \textbf{68.16} & \textbf{52.13} & \textbf{71.73} & \textbf{36.74} & \textbf{50.02} \\
\midrule
\multirow{2}{*}{\tabincell{l}{Semi-\\Supervised}}
& PseudoReasoner \scriptsize{BERT-base} & 71.93 & 84.23 & 66.67 & 63.43 &  45.47 & 68.67 & 30.17 & 41.77 \\
& PseudoReasoner \scriptsize{RoBERTa-L} & \textbf{74.33} & \textbf{85.57} & \textbf{69.33} & \underline{66.37} &  {46.63} & {69.70} & 30.87 & {43.13}  \\
\midrule
Human & & 94.1 & 94.9 & 91.4 & 94.5 & 91.5 & 94.3 & 86.9 & 91.5 \\

\bottomrule
\end{tabular}
\vspace{-0.05in}
\caption{Main experimental results on \ckbpnew{}. Both AUC and F1 are used as evaluation metrics. The ``all'' column indicates the overall performance, and ID, OOD, \textit{Adv.} indicate the performance of the In-domain, Out-of-domain, and Adversarial subset. The best results are \textbf{boldfaced}, and the second-best ones are \underline{underlined}.}
\label{table:overall_result}
\vspace{-0.1in}
\end{table*}


\subsection{Setup}
We examine several models which were previously evaluated on \ckbpold{}, including zero-shot GPT models~\cite{radford2019language}, supervised-learning baselines KG-BERT~\cite{yao2019kgbert} and COMET~\cite{bosselut2019comet}, and semi-supervised-learning models PseudoReasoner~\cite{fang-etal-2022-pseudoreasoner} with two backbone encoders, BERT-base-uncased~\cite{DBLP:conf/naacl/DevlinCLT19} and RoBERTa-large~\cite{DBLP:journals/corr/abs-1907-11692}.
We use Huggingface\footnote{https://huggingface.co/} Transformers~\cite{DBLP:conf/emnlp/WolfDSCDMCRLFDS20} to build our code base. For discriminative models, 
we set the learning rate as 1e-5, batch size 64/32 for base/large variants, respectively, and the number of training epochs as 1. 
For generative models (COMET), we use learning rate 1e-5 and batch size 32 to train in 3 epochs. Negative perplexity scores are used as the final prediction scores.
For PseudoReasoner, we adopt the best settings in \citet{fang-etal-2022-pseudoreasoner}, where we first finetune the KG-BERT model on pseudo-labeling data for one epoch, then from the best checkpoint, we resume the finetuning process on the original training data. 
Note that the training data and unlabeled data are taken from \citet{fang-etal-2022-pseudoreasoner}. 
We run each baseline three times with different random seeds, then average the result and report in Table \ref{table:overall_result}.
For GPT3~\cite{brown2020language} and ChatGPT experiments, we use simple prompts asking them to decide whether an assertion is plausible or not.  

\subsection{Result and Analysis}

The results are shown in Table \ref{table:overall_result}. We provide the AUC score and F1 score of all the baselines on the test set in terms of overall performance (all), performance on the subset of ID, OOD, and \textit{Adv.} samples. 
When calculating F1, for discriminative models, we set the decision threshold as 0.5 (as default), while for generative models, as perplexity serves as the final prediction score, we tune the threshold to obtain the highest F1 score on the development set for each run.

In the zero-shot setting, the scores increase by the version of GPT. GPT3 \texttt{text-davinci-003} gives a significant improvement over GPT2 models, and ChatGPT surpasses its sibling \texttt{text-davinci-003} with a similar margin of improvement. 
Nonetheless, despite the performance improvement from ChatGPT, there is still a clear gap between the zero-shot and (semi-)supervised settings.

In terms of supervised and semi-supervised learning, we observe different scenarios between KG-BERT's performance and COMET's performance, comparing to the result on \ckbpold{} reported in \citet{fang-etal-2022-pseudoreasoner}. 
Here, on \ckbpnew{}, KG-BERT outperforms COMET with a significant gap of 3 AUC overall and also outperforms in all subsets of the test set. This shows the importance of including negative (implausible) examples in the training for discriminating commonsense. This also explains why there is no significant improvement of PseudoReasoner over the baseline KG-BERT on this new evaluation set. 

\begin{figure}[t]
    \centering
    \includegraphics[width=0.8\linewidth]{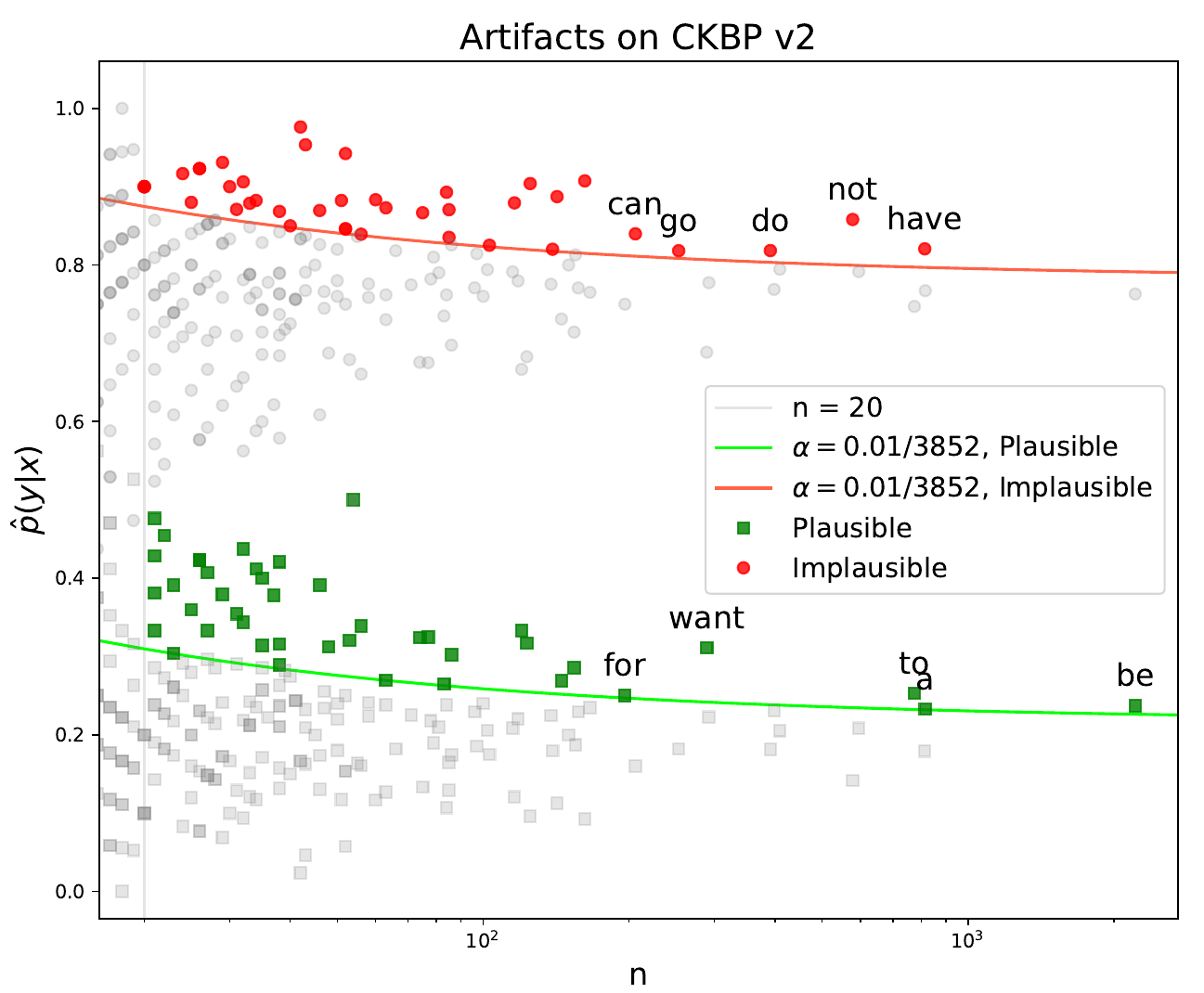}
    \vspace{-0.1in}
    \caption{Artifacts statistics of \ckbpnew{}. Colored dots (either square or circle) represent artifacts in the new evaluation set.}
    \label{fig:artifacts}
    \vspace{-0.1in}
\end{figure}

\subsection{Artifacts Analysis}
There is an uprising acknowledgment of ``artifacts''~\cite{gururangan-etal-2018-annotation, poliak-etal-2018-hypothesis, gardner-etal-2021-competency} in a dataset, in other words, spurious correlations or confounding factors between the surface properties of textual instances and their labels, that may incidentally appear in the annotation process. ``Artifacts'' may undermine the designated evaluation purpose of the dataset. Thus, it is necessary for us to check if ``artifacts'' exist in \ckbpnew{}.

We identify artifacts in \ckbpnew{} by following the previous work ~\citet{gardner-etal-2021-competency}. Particularly, for each word $x$ in the vocab list\footnote{We exclude all relation tokens, as well as special pronoun tokens, namely PersonX, PersonY, PersonZ, PeopleX}, we compute all quantities appearing in the $z$-statistic formula $$z = \dfrac{\hat{p}(y|x) - p_0}{\sqrt{p_0(1-p_0)/n}}.$$
These include word count $n$, estimated probability $\hat{p}(y|x)$ as the fraction of the number of target label $y$ in the corresponding $n$ samples over $n$. After that, we compute the $z$-statistic and reject or not reject the null hypothesis $\hat{p}(y|x) = p_0$ with a significance level $\alpha = 0.01$ and a conservative Bonferroni correction ~\cite{bonferroni1936teoria} for all 3852 vocabulary items. Note that the ``true'' probability $p_0 = p(y|x)$ is taken to be the proportion of samples with label $y$ in the whole evaluation set. Also, we do not consider artifacts with a word count less than 20, as they are not statistically significant.

Figure \ref{fig:artifacts} shows the plot of word count against the estimated probability $\hat{p}(y|x)$ for \ckbpnew{}. 
The additional green and red curves correspond to the largest value of $\hat{p}(y|x)$ w.r.t $n$ to keep the null hypothesis from being rejected, where $y$ takes value ``Plausible'' and ``Implausible'' respectively. 
This means that any dot above the corresponding curve with a frequency of at least 20 is marked as an artifact. 
The artifacts with the largest word count are labeled in the plot. Overall, \ckbpnew{} contains relatively few artifacts (83 artifacts out of 3852 vocabulary items), and the artifacts do not significantly affect the evaluation set quality as their frequencies are not high.

\section{Extrinsic Evaluation}

In this section, we study two downstream applications of CKBP. 
After acquiring a population model, it act as a scoring function to determine whether a triple from the candidate knowledge base $G$ is plausible or not, thus serving as a source of commonsense knowledge acquisition~\cite{DBLP:conf/www/FangZWSH21}.
We leverage the populated knowledge as additional training data for 
both generative commonsense inference (COMET; \citealp{bosselut2019comet}) and zero-shot commonsense question answering \cite{DBLP:conf/aaai/MaIFBNO21}.

\begin{table*}[t]
\small
\centering
\begin{tabular}{@{}l|c|c|c|c|c|c|c}
\toprule
Training Data & BLEU-1 & BLEU-2 & BLEU-3 & BLEU-4 & METEOR & ROUGE-L & CIDEr \\ 
\midrule
ATOMIC &  41.8 & 26.6 & 19.2 & 14.5 & 50.0 & 21.2 & 66.1 \\ 
ATOMIC + CKBP\scriptsize{RoBERTa-L (V1)} & 41.9 & 26.6 & 18.8 & 13.8 & 49.7 & 21.2 & 66.2 \\
ATOMIC + CKBP\scriptsize{RoBERTa-L (V2)} & 42.5 & 26.7 & 18.8 & 13.8 & 50.2 & 21.4 & 67.1 \\
ATOMIC + CKBP\scriptsize{vera} & 42.9 & 27.2 & 19.4 & 14.4 & 50.2 & 21.4 & \textbf{67.5} \\
ATOMIC + CKBP\scriptsize{vera} (mix) & \textbf{43.3} & \textbf{27.6} & \textbf{19.7} & \textbf{14.7} & \textbf{50.3} & \textbf{21.5} & 67.4 \\
\bottomrule
\end{tabular}
\vspace{-0.05in}
\caption{Performance (\%) of GPT2-Large on generative commonsense inference modeling (COMET). 
ATOMIC stands for ATOMIC$_{20}^{20}$ training set, and CKBP stands for our CKBP data. Subscripts under CKBP indicating the population model to select populated commonsense knowledge. The best performances are \textbf{bold-faced}.}
\label{tab:comet_performance}
\vspace{-0.1in}
\end{table*}

\subsection{Generative Commonsense Inference (COMET)}\label{sec:comet}

\paragraph{Setup}

We follow the basic settings as in the original  ATOMIC$_{20}^{20}$ paper~\cite{hwang2021comet} to generate commonsense tails $t$ given head $h$ and relation $r$ as input.
The evaluation dataset is the annotated 5,000 test examples provided by \citet{hwang2021comet}.
We use BLEU~\cite{DBLP:conf/acl/PapineniRWZ02}, ROUGE-L~\cite{lin-2004-rouge}, METEOR~\cite{DBLP:conf/wmt/LavieA07}, and CIDEr~\cite{DBLP:conf/cvpr/VedantamZP15} as the automatic evaluation metrics. 
 

Specifically, we compare the performance of the following training paradigms: 1) Training the model using the official training set of ATOMIC$_{20}^{20}$. 2) Pre-training the model using a comparable amount of CKBP-acquired data, and subsequently fine-tune on ATOMIC$_{20}^{20}$ training set.  3) Training on a mixture of CKBP-acquired data and ATOMIC$_{20}^{20}$ training data.

We filter the CKBP-acquired data using two filters.
First, we employ two typical population models, RoBERTa-L~\cite{DBLP:journals/corr/abs-1907-11692} fine-tuned on CKBP training set and Vera ~\cite{liu2023vera} to provide a plausibility score for each triple. 
We set an empirical threshold of 0.8 and selecting triples with plausibility score higher than that as populated commonsense knowledge.
For the RoBERTa-L model, we select the best-performed checkpoints based on both \ckbpold{} and \ckbpnew{} to evaluate which evaluation set is better aligned with downstream performance.
Second, we utilize a diversity filter defined in G-DAUG~\cite{DBLP:conf/emnlp/YangMFSBWBCD20}, which is a heuristic favoring diverse n-grams. 
The diversity filter is applied such that we select the same amount of CKBP-acquired data as the training set of ATOMIC$_{20}^{20}$.

We choose GPT2-Large as our backbone language model. We didn't use GPT2-XL as in \citet{hwang2021comet} because the XL version performs relatively poorer than the Large version in terms of most automatic evaluation metrics on the evaluation set of ATOMIC$_{20}^{20}$ despite twice the model size.
The learning rate is set as 1e-5, and we train the model for three epochs on both CKBP-acquired data and ATOMIC$_{20}^{20}$ training data.

\paragraph{Results and Analysis}
The results of generative commonsense inference are presented in Table \ref{tab:comet_performance}.
First, adding CKBP-acquired commonsense knowledge for either pre-training or co-training can yield a general performance improvement in generative commonsense inference. Specifically, the model trained on ATOMIC + CKBP $\text{\scriptsize{Vera}}$ 
achieves the best performance and outperforms that only fine-tuned on ATOMIC$_{20}^{20}$ on all automatic evaluation metrics.
This indicates that leveraging the abundant unlabeled discourse knowledge from ASER ($G$), accompanied by appropriate plausibility filtering through the population model, can effectively serve as valuable augmented data to enhance commonsense reasoning.
Among the population models, we observe that a better population model, as evaluated by our \ckbpnew{} evaluation set, corresponds to a higher performance gain in the generative commonsense inference task. This finding highlights the promising potential of developing improved population models, which subsequently contribute to enhanced downstream applications.


Second, the RoBERTa-L model selected by \ckbpnew{} demonstrates greater efficacy in enhancing generative commonsense inference compared to the model selected by \ckbpold{}. This finding suggests that \ckbpnew{} exhibits improved alignment with real-world downstream applications, surpassing its predecessor in terms of practical utility.
It's also noteworthy that COMET is an important task that inherently benefits a pile of further downstream tasks that requires commonsense reasoning, including zero-shot commonsense question answering with self-talk~\cite{DBLP:conf/emnlp/ShwartzWBBC20} and dynamic graph construction~\cite{DBLP:conf/aaai/BosselutBC21}, narrative reasoning~\cite{peng-etal-2022-inferring}, and dialogue generation~\cite{tu-etal-2022-misc}. 
In this regard, our work exhibits significant potential for generalization to tasks extending beyond the realm of commonsense reasoning.

\begin{table*}[t]
\small
\centering
\begin{tabular}{@{}l|c|lllll|l@{}}
\toprule
Model & CSKB & a-NLI & CSQA & PIQA & SIQA & WG & Avg. \\ 
\midrule
\multicolumn{8}{@{}l}{\textbf{Zero-shot Baselines}} \\
Random & - & 50.0 & 20.0 & 50.0 & 33.3 & 50.0 & 40.7 \\
Majority & - & 50.8 & 20.9 & 50.5 & 33.6 & 50.4 & 41.2 \\
RoBERTa-L~\cite{DBLP:journals/corr/abs-1907-11692} & - & 65.5 & 45.0 & 67.6 & 47.3 & 57.5 & 56.6 \\
DeBERTa-v3-L~\cite{he2023debertav} & - & 59.9 & 25.4 & 44.8 & 47.8 & 50.3 & 45.6 \\
Self-talk~\cite{DBLP:conf/emnlp/ShwartzWBBC20} & - & - & 32.4 & 70.2 & 46.2 & 54.7 & - \\
COMET-DynGen~\cite{DBLP:conf/aaai/BosselutBC21} & ATOMIC & - & - & - & 50.1 & - & - \\
SMLM~\cite{DBLP:conf/emnlp/BanerjeeB20} & * & 65.3 & 38.8 & - & 48.5 & - & - \\
MICO~\cite{DBLP:conf/emnlp/SuWFZSZ22} & ATOMIC & - & 44.2 & - & 56.0 & - & - \\
STL-Adapter~\cite{DBLP:conf/naacl/KimKKAHY22} & ATOMIC & 71.3 & 66.5 & 71.1 & 64.4 & 60.3 & 66.7 \\
\midrule
\multicolumn{8}{@{}l}{\textbf{Backbone: DeBERTa-v3-Large} \scriptsize{\textit{435M}}} \\
DeBERTa-v3-L (MR)~\cite{DBLP:conf/aaai/MaIFBNO21} & ATM-10X & 75.1 & \underline{71.6} & \textbf{79.0} & 59.7 & 71.7 & 71.4 \\
DeBERTa-v3-L (MR)~\cite{DBLP:conf/aaai/MaIFBNO21} & ATOMIC & 76.0 & 67.0 & \underline{78.0} & 62.1 & \underline{76.0} & \underline{71.8} \\
DeBERTa-v3-L (MR)~\cite{DBLP:conf/aaai/MaIFBNO21} & CKBP (our) & \textbf{79.2} & 69.6 & 77.9 & 64.3 & \textbf{77.2} & \textbf{73.6} \\
\midrule
\multicolumn{8}{@{}l}{\textbf{Large Language Models}} \\
GPT-3.5 (\texttt{text-davinci-003}) & - & 61.8 & 68.9 & 67.8 & \underline{68.0} & 60.7 & 65.4 \\
ChatGPT (\texttt{gpt-3.5-turbo}) & - & 69.3 & \textbf{74.5} & 75.1 & \textbf{69.5} & 62.8 & 70.2 \\
\midrule
\rowcolor{gray!30} \multicolumn{8}{@{}l}{\textbf{Supervised Learning \& Human Performance}} \\
\rowcolor{gray!20} RoBERTa-L (Supervised) & - & 85.6 & 78.5 & 79.2 & 76.6 & 79.3 & 79.8 \\
\rowcolor{gray!20}DeBERTa-v3-L (Supervised) & - & 89.0 & 82.1 & 84.5 & 80.1 & 84.1 & 84.0 \\
\rowcolor{gray!20}Human Performance & - & 91.4 & 88.9 & 94.9 & 86.9 & 94.1 & 91.2 \\
\bottomrule
\end{tabular}
\vspace{-0.1in}
\caption{Zero-shot evaluation results (\%) on five commonsense question answering benchmarks. 
The best results are \textbf{bold-faced}, and the second-best ones are \underline{underlined}.
The performance of supervised learning and human are for reference only.
}
\label{tab:main_exp_results}
\vspace{-0.10in}
\end{table*}

\subsection{Zero-shot Commonsense QA}

\paragraph{Setup}
For the zero-shot commonsense question answering (QA) task, we adopt the task definition and evaluation pipeline proposed by~\citet{DBLP:conf/aaai/MaIFBNO21} to evaluate the benefit \ckbpnew{} brings to extrinsic QA. 
Several methods have been proposed to tackle this task, including those by~\citet{DBLP:conf/emnlp/ShwartzWBBC20,DBLP:conf/aaai/BosselutBC21,DBLP:conf/naacl/KimKKAHY22}
The most effective pipeline, as proposed by~\citet{DBLP:conf/aaai/MaIFBNO21}, injects commonsense knowledge into pre-trained language models through fine-tuning on QA pairs synthesized from knowledge in CSKBs.
To perform this fine-tuning, the head $h$ and relation $r$ of a $(h, r, t)$ triple are transformed into a question using natural language prompts, while the tail $t$ is used as the correct answer option. 
Distractors or negative examples are created by randomly sampling tails from triples that do not share common keywords with the head. 
This fine-tuning process enhances the model's knowledge not only for QA benchmarks constructed from CSKBs, such as SocialIQA~\cite{DBLP:conf/emnlp/SapRCBC19} derived from ATOMIC, but also improves its ability to answer previously unseen commonsense questions in a more generalized manner.

We adopt the original QA synthesis and model training pipeline by~\citet{DBLP:conf/aaai/MaIFBNO21} on the original ATOMIC and the one augmented with populated knowledge from~\ckbpnew{} to ablatively study the sole benefit that knowledge in~\ckbpnew{} brings.
Similar with that in COMET experiments, we use the best-performed CKBP model, Vera, to score the whole population space in ASER and select the populated knowledge with plausibility scores of over 0.8. 
Then the same diversity filter as in Section~\ref{sec:comet} is used to downsample the number of populated triples to be comparable with the size of the training set in ATOMIC$_{20}^{20}$.
For the QA model, DeBERTa-v3-Large~\cite{he2023debertav} is used as the backbone, and we train the model using a learning rate of 7e-6 for one epochs on both the CKBP-acquired data and ATOMIC-synthesized data as provided by \citet{DBLP:conf/aaai/MaIFBNO21}.

Once trained, we evaluate the model on the validation splits of five commonsense QA benchmarks: Abductive NLI (aNLI;~\citealp{DBLP:conf/iclr/BhagavatulaBMSH20}), CommonsenseQA (CSQA;~\citealp{DBLP:conf/naacl/TalmorHLB19}), PhysicalIQA (PIQA;~\citealp{DBLP:conf/aaai/BiskZLGC20}), SocialIQA (SIQA;~\citealp{DBLP:conf/emnlp/SapRCBC19}), and WinoGrande (WG;~\citealp{DBLP:journals/cacm/SakaguchiBBC21}).
Accuracy is used as the evaluation metric.
Furthermore, we compare our model not only against existing zero-shot knowledge injection methods~\cite{DBLP:conf/emnlp/ShwartzWBBC20,DBLP:conf/aaai/BosselutBC21,DBLP:conf/emnlp/BanerjeeB20,DBLP:conf/emnlp/SuWFZSZ22,DBLP:conf/naacl/KimKKAHY22,DBLP:conf/aaai/MaIFBNO21} but also against large language models such as ChatGPT~\cite{openai2022chatgpt} and GPT-3.5~\cite{DBLP:conf/nips/BrownMRSKDNSSAA20}.

\paragraph{Results and Analysis}
The zero-shot commonsense QA results are shown in Table~\ref{tab:main_exp_results}.
Among all the zero-shot methods, the model trained on~\ckbpnew{} demonstrates the highest performance. 
It outperforms models trained solely on ATOMIC (with an increase of 2.2\%) and ATOMIC10X~\cite{DBLP:conf/naacl/WestBHHJBLWC22} (with an increase of 1.8\%). 
Importantly, our method surpasses large language models by an average of 3.4\%. 
This performance gain highlights the significant advantage of our populated commonsense knowledge over both human annotations and distilled knowledge from large language models.
Furthermore, we observe that the model trained on CKBP-acquired data shows the most improvement on the aNLI and WinoGrande benchmarks. 
One potential reason for this is that the populated knowledge in~\ckbpold{} encompasses a wider range of commonsense knowledge beyond only social commonsense, which benefits tasks involving abductive reasoning (based on narrative) and pronoun coreference resolution.

\section{Conclusion}
In this paper, we introduce a new CSKB Population benchmark \ckbpnew{} which addresses two problems of the predecessor \ckbpold{}. Besides, we conduct a broad range of experiments with different models, including GPT3.5 and ChatGPT, on the new evaluation set. The result shows that the CSKB Population task remains a hard task of commonsense reasoning even for state-of-the-art LLMs, which challenges the community for future research.


\section*{Limitations}

We observe several limitations of this work. First, \ckbpnew{} still follows the lemmatized format of events, which may hinder the usage of the resulting population model on knowledge bases other than ASER. 
Second, the paradigm of CSKB is context-free, which may have difficulty in directly applying to actual downstream tasks.
Third, As this paper focuses on proposing a new evaluation set of the CSKB Population, we do not present novel tailored methods for solving this task, leaving it to future research.

\section*{Ethical Statements}
This work presents \ckbpnew{}, an open-source benchmark for the research community to study the CSKB population problem.
The training set is directly adapted from \ckbpold{} and ATOMIC($_{20}^{20}$), GLUCOSE, and ConceptNet, which would have the same ethical issues as in those previous works.
Instances in the evaluation set are retrieved from \ckbpold{} and ASER, both being open-source with an MIT license.
Events in all data instances are anonymized. 
Thus, the benchmark does not pose any privacy problems about any specific entities (e.g., a person or company). 
We carried out human expert annotation, where annotators are fairly paid according to the minimum wage requirement of the local government.

\bibliography{custom}




\end{document}